\documentclass[conference,a4paper]{ieeetran}
\usepackage{algorithm,algcompatible}
\usepackage{algpseudocode} 
\usepackage{graphicx, color}
\usepackage{cite}
\graphicspath{ {images/} }
\usepackage{amsmath}
\usepackage[table]{xcolor}
\usepackage{pbox}
\usepackage{caption}
\usepackage{multirow}
\usepackage{amssymb}
\usepackage{booktabs,flushend}

\usepackage[bookmarks=false]{hyperref}

\usepackage{fancyheadings}
\title{Learning Autoencoded Radon Projections}

\author{
 \authorblockN{ Aditya Sriram\authorrefmark{1}, Shivam Kalra\authorrefmark{1}, H.R. Tizhoosh\authorrefmark{1}, Shahryar Rahnamayan\authorrefmark{2}
}
\authorblockA{\authorrefmark{1} KIMIA Lab, University of Waterloo, Canada}
\authorblockA{\authorrefmark{2} Elect., Comp. and Software Eng., University of Ontario Institute of Technology, Canada
}}

\begin{document}
\maketitle
\pagenumbering{gobble}
\begin{abstract}
Autoencoders have been recently used for encoding medical images. In this study, we design and validate a new framework for retrieving medical images by classifying Radon projections, compressed in the deepest layer of an autoencoder. As the autoencoder reduces the dimensionality, a multilayer perceptron (MLP) can be employed to classify the images. The integration of MLP promotes a rather shallow learning architecture which makes the training faster. We conducted a comparative study to examine the capabilities of autoencoders for different inputs such as raw images, Histogram of Oriented Gradients (HOG) and normalized Radon projections. Our framework is benchmarked on IRMA dataset containing $14,410$ x-ray images distributed across $57$ different classes. Experiments show an IRMA error of $313$ (equivalent to $\approx 82\%$ accuracy) outperforming state-of-the-art works on retrieval from IRMA dataset using autoencoders.   
\end{abstract}

\section{Introduction}
\label{sec:introduction} 
Computer-aided diagnosis (CAD) systems are designed to assist medical specialists by interpreting medical images and classifying them based on descriptive semantics extracted from each image \cite{doi2007computer}  \cite{fujita2007computer}. Conventionally, medical images are evaluated by specialists; however, interpretations by humans are understandably subject to errors due to intricate anatomical vagueness, inter-observer variability, and tediousness of the task. Content-based image retrieval (CBIR) can be a crucial part of next-generation CAD systems by serving as ``virtual peer review''; retrieving and displaying similar cases along with their archived reports can help the clinicians to make more reliable decisions.  

Recent developments in machine learning have opened up new opportunities for image retrieval. Among others, \emph{autoencoders},  introduced in 2006 by Hinton and Salakhutdinov \cite{hinton2006reducing}, are designed to compress high-dimensional data by removing redundancies and preserving salient features. Deng in \cite{deng2014tutorial} presented a notable survey on autoencoders, covering the mathematical aspects and a general overview of the types of learning and applications for which autoencoders can excel. Goyal and Benjamin \cite{goyal2014object}, on the other hand, describe the history of deep learning and describe various image data sets for which the deep learning architectures can be exploited. 

Initially, autoencoders were trained on the MNIST dataset \cite{hinton2006reducing} for extracting features and retrieval \cite{krizhevsky2011using}. In 2015, Camlica et al. \cite{camlica2015autoencoding} presented an autoencoder to eliminate image regions that possess low encoding errors by validating its retrieval accuracy against both local binary patterns (LBP) and support vector machines (SVM). Tested against the IRMA dataset, roughly $50$\% of the image area was removed which increased the retrieval speed by $27$\% with less than $1$\% decrease in the accuracy. Sze-To et al.  \cite{sze2016binary} developed a Radon Autoencoder Barcode (RABC) to hash images into binary codes using the IRMA dataset. Their method obtained the lowest total error of $\approx344$ using 512-bit codes for retrieval. In 2016, Tizhoosh et al. \cite{tizhoosh2016barcodes}, introduced Autoencoded Radon Barcode (ARBC) to autoencode Radon projections using mini-batch stochastic gradient descent by binarizing the outputs from each hidden layer during training, and to produce a \emph{barcode} per layer. A comparison with other methods (e.g., RBC, SURF, and BRISK) suggested that ARBC can achieve the lowest  IRMA error of $\approx392$. 

Motivated to exploit the potentials of autoencoders for medical image retrieval, in this work we propose a novel approach which classifies autoencoded Radon projections using MLPs to calculate the class probability for each query image. Radon transform is widely used in medical applications as means for reconstructing images using   equidistant parallel rays penetrating an object from many directions \cite{sanz2013radon} \cite{deans2007radon}. Specifically, an autoencoder is adopted for dimensionality reduction. The autoencoded features (compressed image/histogram/projections) are then classified by an MLP to assign the image to a class. We examined Radon projections as features based on some recent success; Shujin and Tizhoosh \cite{zhu2016radon} combined Radon projections and support vector machines (SVM) for CBIR tasks. As well, Tizhoosh et. al \cite{tizhoosh2016minmax} presented MinMax Radon barcodes which are observed to retrieve images $15$\% faster compared to ``local thresholding'' \cite{Tizhoosh2015rbc}. 

In order to evaluate the algorithm, the IRMA repository containing $14,410$ x-ray images along with their labels called ``IRMA code'', are used. The proposed method is implemented using \emph{Keras} and \emph{Theano}, and the networks were trained on Nvidia's Tesla K-80 GPUs provided by SHARCNET \cite{sharc2017}.


\section{The Proposed Method}
\label{sec:methodology}

\begin{figure*}[htb]
  \centering
  \includegraphics[width=\textwidth]{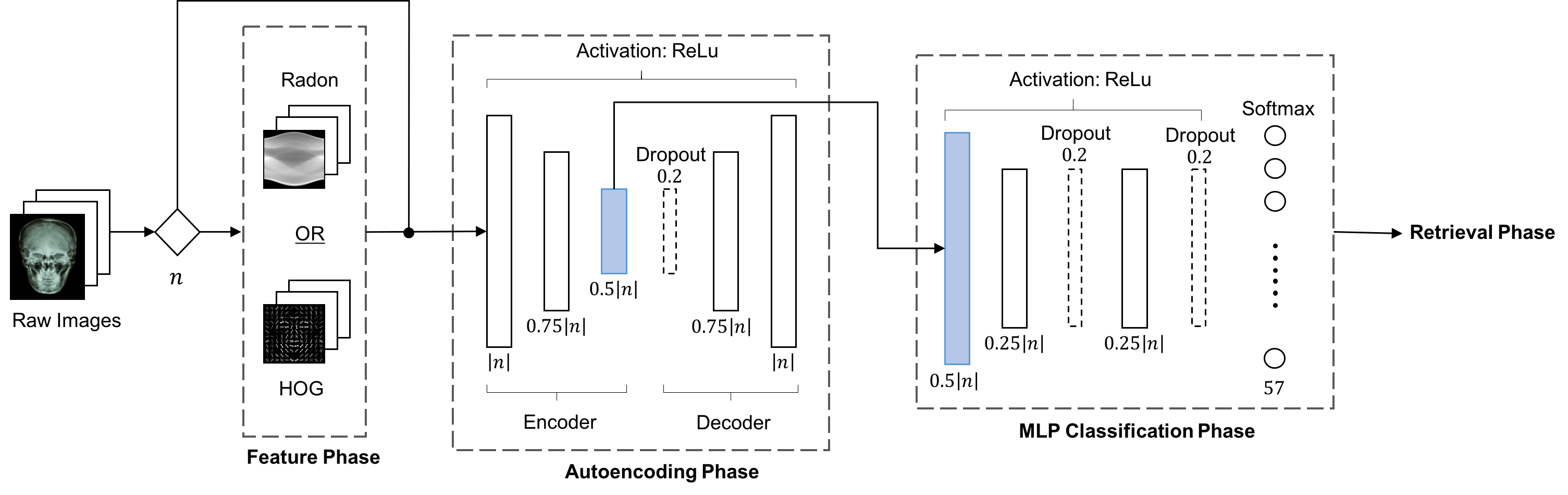}
	\caption{Classification of x-ray images into $57$ classes by MLP using
    the deepest layer of autoencoder trained on raw pixels, HOG or Radon projections.}
  \label{fig:classification}
\end{figure*}

The proposed method is comprised of three components: (\textit{i}) dimensionality reduction, (\textit{ii}) classification, and (\textit{iii}) retrieval. To begin, Radon projections are extracted from each image and vectorized to form an input to the autoencoder which should remove redundancies by compressing the image while preserving salient features. The classification component, an MLP, establishes a feature-to-category relationship, wherein features are compressed Radon projections in the deepest autoencoded layer. Fig. \ref{fig:classification} provides a schematic representation of the proposed approach. For sake of comparison, we trained the autoencoder with raw pixels, HOG features, and Radon projections to find most descriptive inputs for the autoencoder.

The proposed architecture is divided into three phases: (\textit{i}) feature extraction, (\textit{ii}) training of the autoencoder, and (\textit{iii}) using autoencoder's deepest layer as features for training the MLP classifier. There are three types of features (Radon, HOG, and raw images) that are extracted and pre-processed prior to training the autoencoder. The pre-processing steps depend on the type of input feature provided to the network. Raw images are rescaled between $0$ and $1$, and padded to maintain a square size. Radon projections, on the other hand, are normalized by dividing each projection by the maximum value, and then standardized, with $\mu=0, \ \sigma=1$. Finally, HOG features are provided to the network unprocessed. Each of the raw features are vectorized, and provided to the autoencoder for compression. To obtain an optimally trained network, early stopping along with 10-fold cross validation is computed for both autoencoder and MLP networks. The final performance measure is the average of the values computed in the loop of the 10-folds -- which ensures the creation of an optimal model when the number of samples is very small. The final classification probabilities are produced by the best configuration. 


There are several motivations behind the multi-phase training architecture for the proposed method. The deep features from the autoencoder are observed to preserve salient attributes that are free from redundancies; this property enhances the convergence and classification accuracy of  MLP. It also enables us to work with a rather shallow network that is easier to train. Additionally, the IRMA dataset is a rather small archive, therefore, training autoencoder and MLP separately promises better convergence. Moreover, a combination of these networks could further be fine-tuned as one network, hence a two-phase training can be thought of as pre-training steps for both autoencoder and MLP. For testing the network, each query is passed through both trained autoencoder and the MLP  to obtain a final class assignment which is used later for similarity matching during the retrieval process.

\begin{algorithm}
  \caption{Preprocessing the image $I$ and its features $I_f$}
  \label{algo:preprocess}
  \begin{algorithmic}[1]

    \Procedure{GetFeatures}{$I,\ F_s$} 
    \State $I_{p} \gets double(I)$ (Normalize to $ [0,1]$)
    \State $D_{\max} \gets \max(I_p.shape)$ (Find max dimension of the image)
    \State $I_{p} \gets Pad(I_p, D_{\max})$ (Convert $I_p$ to
      $D_{\max} \times D_{\max}$ size, aligning to center of larger dimension)
    \State $I_{p} \gets Resize(I_p, (N,N))$ (resizes $I_p$ to $N \times N$)
    \If{$F_S$ is HOG} 
    \State{$I_f\gets\ ReShape2Vector(HOG(I_q, n_{hog}))$} (reshapes 2D to 1D)
    \ElsIf{$F_s$ is Radon}
    \State{$I_f\gets\ Normalize(R(I_q, \theta), axis=1)$}
    \State{$I_f \gets ReShape2Vector(I_f)$}
    \Else
    \State{$I_f\gets\ ReShape2Vector(I_q)$}
    \EndIf{}
    \Return{$I_f$}
    \EndProcedure{}
    
  \end{algorithmic}
\end{algorithm}

\begin{figure*}[tb]
  \centering
  \includegraphics[width=0.9\textwidth]{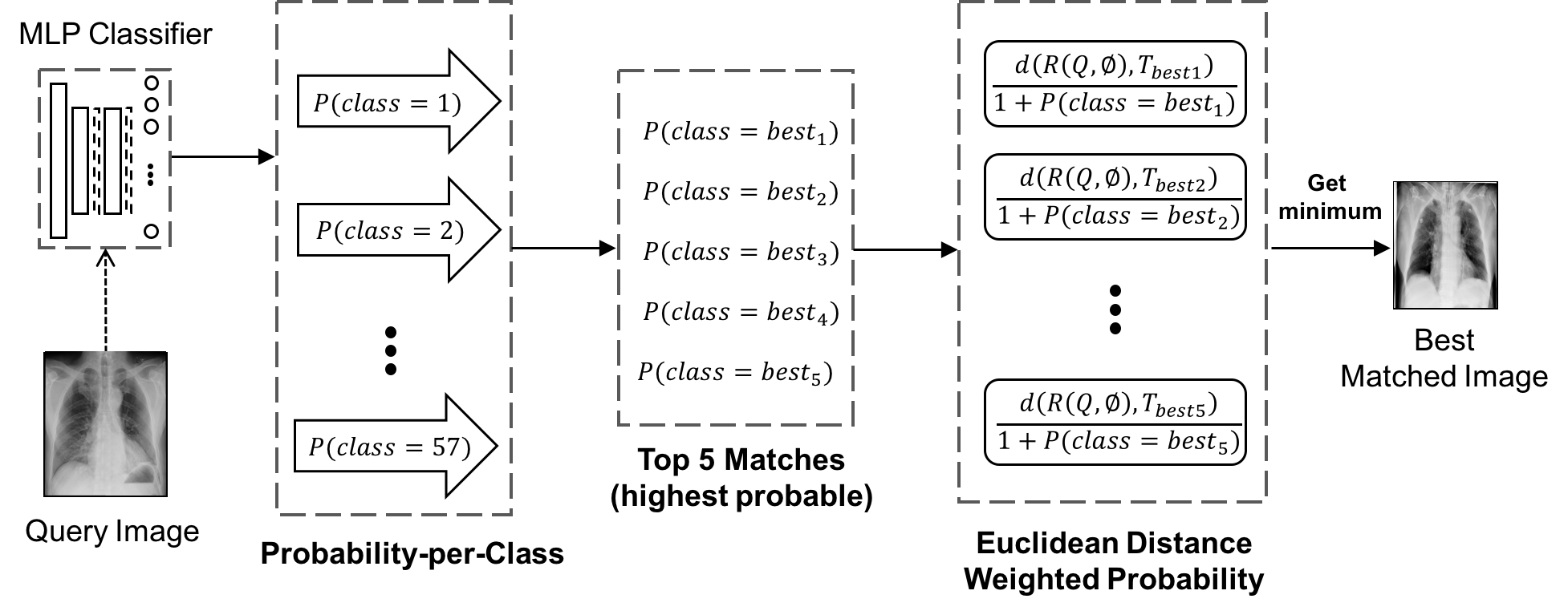}
  \caption{Image retrieval using KNN search within the top 5 predicted classes.}
  \label{fig:retrieval}
\end{figure*}

For retrieving similar images, every query image is reshaped and processed to generate an input vector for the autoencoder (see Algorithm \ref{algo:preprocess}). Since the proposed architecture is sequential, the query features from the deepest layer of the autoencoder are transferred into MLP for classification. According to the predicted image class label, determined by the MLP, the images within the top-five predicted categories are selected from the database. Thereafter, the equidistant normalized Radon projections generated from the query image are compared with the top $5$ images by using the $k$-nearest neighbor method, yielding the index of the most similar image. For instance, for Radon projections as input features, the top $5$ probability classes are determined using the MLP after autoencoding the features. To retrieve the most similar image, the query image along with the top $5$ probable indexed images are transformed into Radon projections. Thereafter, Euclidean distance between the projections is computed to retrieve the best matched image. In essence, the MLP is employed for global search (class assignment), and the input feature is computed for local search (to retrieve the most similar image). Analogously, the same procedure is applied to HOG and raw images, when they are provided as input features to the designed framework (see Algorithm \ref{algo:preprocess}). The overall architecture is illustrated in Fig. \ref{fig:retrieval} and Algorithm \ref{algo:retrieval}. In essence,  a total of 57 probabilistic outputs are ranked from which the top 5 are used in a KNN search to find the best match for the query image. It is important to note that KNN with $L_2$ distance did provide the best accuracy.
\begin{algorithm}[!htb]
  \caption{Retrieving the best match for the query 
    $I_q$}
  \label{algo:retrieval}
  \begin{algorithmic}[1]
   \State $I_q$: the query
      Image, $AE$: the trained autoencoder, $MLP$: the trained MLP,  $F_S$: defined input features (i.e., Radon, HOG, raw images)
    \Procedure{Retrieve}{$I_q,\ AE, MLP, F_s$} 
    \State{$I_f^q \gets \textrm{GetFeatures}(I_q, F_s)$} 
    \State{$I_\textrm{enc} \gets \textrm{Autoencoder}(I_f)$}
    \State{$P_c \gets \textrm{MLP}(I_\textrm{enc})$} (get class probabilities)
    \State{$\textrm{Top}_{idx}\gets \textrm{argsort}(P_c,\ 'decreasing')[:5]$}
     (get indexes of top five classes)
    \State{$D=[]$} (Initialize empty array)
    \State{$\textrm{Idx}=[]$} 
    \State $m=0$ (number of candidate images)
    \For  {$i$ in $\textrm{Top}_\textrm{idx}$}
      \For {$j$ in $|C_i|$ } (for all images in class $C_i$)
    	\State{$I_f^{i,j} \gets \textrm{GetFeatures}(I_i^j, F_s)$} (get features for the candidate image)
	\State{$d, \textrm{idx} \gets \textrm{KNN}(I_f^{i,j},I_f^q)$} (perform knn search)
	\State $m=m+1$
    	\State{$D[m] \gets \frac{d}{1 + P_c[i]}$} (normalize w.r.t. to prediction probability)
    	\State{$\textrm{Idx}[m] \gets idx$} 
      \EndFor{}	
    \EndFor{}
    \State{$Best\_Match = \textrm{Idx}[\textrm{argmin}(D)]$} \\
    \Return{Best\_Match}
    \EndProcedure{}
  \end{algorithmic}
\end{algorithm} 

\section{Experiments}
\label{sec:experiments}

The Image Retrieval in Medical Applications (IRMA) is a publicly available x-ray data set collected at the Department of Diagnostic Radiology at the RWTH Aachen University, in Germany \cite{irmaWeb}. The repository contains $12,677$ training and $1,733$ testing x-ray images composed of cases of patients of different ages, genders, captured at different angular positions, and depicting various pathologies \cite{tommasi2009overview}. This data set is particularly challenging for machine learning algorithms due to its noisy radiographs and nonuniform distribution that is extremely biased towards the first category, namely the chest radiographs. Each image possesses a unique ``IRMA code'' which belongs to one of the 193 categories (2009 version), or 57 categories (2005 version). For validation, the IRMA codes of the query and retrieved images are compared to provide an error between $0$ and $1$. It is important to note that these IRMA codes are designed by professionals solely for benchmarking purposes, and do not have any relevance in the real world. The total IRMA error is calculated, as a final metric, when comparing each query image to the best-matched train image (least distance between the feature vectors). The summation of the best-matched IRMA errors quantifies the performance the network, which is then compared to literature. 

We performed three series of experiments, one for each feature
input to the autoencoder, i.e., (\textit{i}) Radon projections, (\textit{ii})
HOG features, and (\textit{iii}) raw pixels. For every series of experiments, we alter (\textit{i}) the image size, and (\textit{ii}) feature dependent parameter(s) such as the number of equidistant projections for Radon transform, the number of histograms for HOG features, and compression (downsampling) ratio for raw pixels. It is important to note that the same type of input feature, that is provided to the network, is used for retrieval process. Finally, the quality of the solution is assessed based on the training accuracy, testing accuracy, and the accumulated IRMA error. To retrieve the lowest IRMA error, namely when using Radon projections, $57$ indexed KD-trees are generated; wherein the $i^{th}$ KD-Tree is indexed using $16$ equidistant 1D vectorized and normalized Radon projections from pre-processed $256\times256$ training images which belong to the $i$-the category. The ``best match'' for every query image is found based on the minimum distance when comparing its vector representation against all vectors in the training set. Particularly, $16$ equidistant projections are chosen based on results presented in \cite{tizhoosh2016barcodes}. As well, a $256\times256$ size was chosen empirically to unify the image dimensions to simplify the retrieval task. For the first experiment, the Radon projections are calculated from pre-processed training images and normalized on a per-projection basis. The normalized Radon projections are then concatenated to form a 1D vector which is fed into the autoencoder. This experiment yielded an IRMA error of $313$, which is the best result on the IRMA dataset using an autoencoder. For the second series of experiments, we used HOG descriptors as input to the autoencoder, giving us an IRMA error of $\approx326$ for $8$ histograms. Finally, for the third series of experiments, we used raw images as input to the autoencoder, giving us an IRMA error of $\approx 349$ with 50\% compression using $64\times64$ input images. Note that there is a $25$\% reduction in size per layer, and we tested using 1 or 2 hidden layers, giving $25$\% and $50$\% dimensionality reduction, respectively. Table\ref{tab:table_experiment} shows the experimental results (dark gray cells: the best results; light gray cells: the best value for a given input dimension).

\begin{table*}[tb]
\centering
\caption{Comparative results: For different image sizes (first column), we measured the IRMA error and network accuracies (second column). For Radon projections (third column) four scenarios were tested with 8, 10, 16 and 20 equidistant projections. For HOG features (fourth column), four different gradient histograms were calcutes with 4, 8, 10 and 16 directions. Raw images (last column) were compressed at 25\% and 50\%. }
\begin{tabular}{|c|l|c c c c|c c c c|c c|}
\hline
\multirow{2}{*}{\textbf{\begin{tabular}[c]{@{}c@{}}Input size\end{tabular}}} & \multirow{2}{*}{\textbf{\begin{tabular}[c]{@{}c@{}}Error/Accuracy\end{tabular}}} & \multicolumn{4}{c|}{\textbf{\begin{tabular}[c]{@{}c@{}}Radon\end{tabular}}} & \multicolumn{4}{c|}{\textbf{\begin{tabular}[c]{@{}c@{}} HOG\end{tabular}}} & 
\multicolumn{2}{c|}{\textbf{\begin{tabular}[c]{@{}c@{}}~Raw Image~ \end{tabular}}} \\ \cline{3-12} 
                                                                                           &                                                                                         & $\textbf{8}$                           & $\textbf{10}$                         & $\textbf{16}$                         & $\textbf{20}$   & $\textbf{4}$           & $\textbf{8}$         & $\textbf{10}$         & $\textbf{16}$         & $\textbf{25}$\%                                                  & $\textbf{50}$\%                                                 \\ \hline
$32\times32$                                                                     & IRMA Error                                                                              & $345$                         & $341$                        & \cellcolor[HTML]{D3D3D3}$335$                        & $340$                        & $410$         & $408$        & $408$        & \cellcolor[HTML]{D3D3D3}$405$        & \cellcolor[HTML]{D3D3D3}$351$                                                   & $354$                                                  \\  
$64\times64$                                                                     & IRMA Error                                                                              & $332$                         & $332$                        & $330$                        & \cellcolor[HTML]{D3D3D3}$328$                        & $329$         & \cellcolor[HTML]{A9A9A9}$326$        & $328$        & $331$        & \cellcolor[HTML]{A9A9A9}$349$                                                   & $353$                                                  \\  
$128\times128$                                                                   & IRMA Error                                                                              & $325$                         & $323$                        & $323$                        & \cellcolor[HTML]{D3D3D3}$320$                        & $332$         & \cellcolor[HTML]{D3D3D3}$329$        & $335$        & $335$        & $349$                                                   & \cellcolor[HTML]{D3D3D3}$350$                                                  \\ 
$256\times256$                                                                   & IRMA Error                                                                              & $317$                         & $316$                        & \cellcolor[HTML]{A9A9A9}$313$                        & $314$                        & $336$            &    \cellcolor[HTML]{D3D3D3}$333$         & $341$           & \cellcolor[HTML]{D3D3D3}$350$           & $351$                                                      & $356$                                                     \\ 
 \hline
\end{tabular}
\label{tab:table_experiment}
\end{table*}


\section{Discussion and Conclusion}
\label{sec:discussion}
\begin{figure*}[tb]
  \centering
  \includegraphics[width=0.75\textwidth]{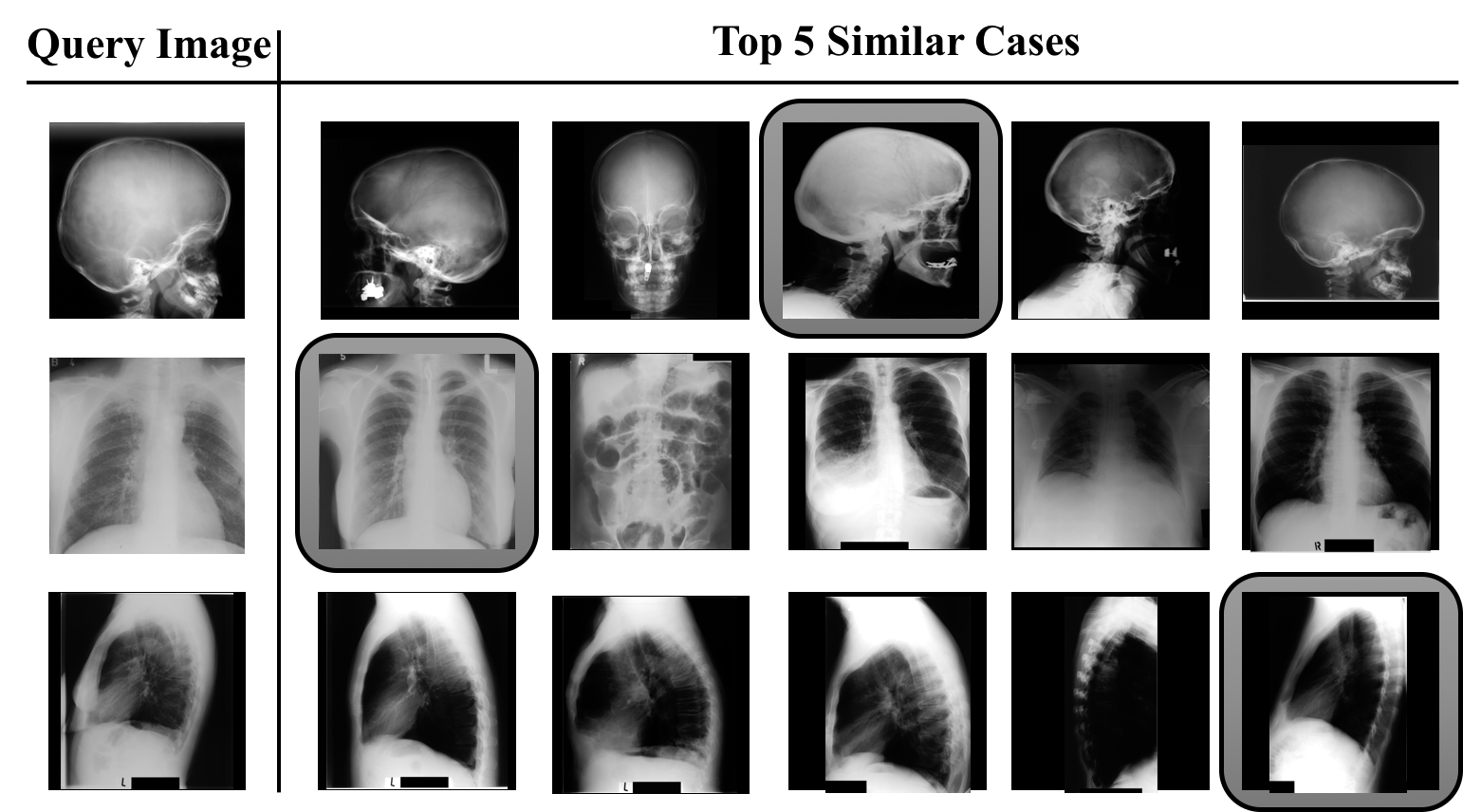}
\caption{Examples for the ``best match'' (highlighted) using Radon projections. Each retrieved image has its own IRMA code which can be compared with the query's IRMA code to calculate the retrieval error. }
  \label{fig:best_match}
\end{figure*}

Investigating the applicability of the autoencoders in medical image retrieval, the lowest IRMA error of $313$ was achieved by training a classifier using $16$ equidistant normalized Radon projections from pre-processed training images with a 10-fold cross validation and early stopping. Radon projections performed better than both the HOG features and raw pixel values. In addition, Radon projections were observed to consume less memory and to be faster ($\approx$ 25GB of RAM for 2.5 hours) when compared to HOG ($\approx$ 30GB of RAM for 3 hours) and raw images ($\approx$ 35GB of RAM for 4.5 hours). Additionally, weighting the retrieval distances using the MLP classification probability is observed to consistently yield better results compared to simply using Euclidian distances. Fig. \ref{fig:best_match} shows the effect of weighting of similarity using classification probabilities. The images from left to right are sorted in increasing order of Radon projection similarity, wherein the highlighted image shows the best match after probability weighting. 

We also augmented the data set to obtain $70,000$ images. However, despite the increase during the training, the IRMA error was similar to the best achieved. Moreover, after several experiments on various other learning parameters, the best activation function was observed to be \emph{ReLU} (rectified linear unit) when compared to \emph{Sigmoid}, \emph{tanh}, and \emph{rmsprop} functions. We also tried different values of probabilities for the Dropout layer, and the best value achieved from our test was $0.2$ for all models. Table \ref{tab:IRMAresults} shows the relative ranking of the proposed approach against existing methods in the literature. Beside the IRMA error ($E_\textrm{tot}\in [0,1733]$), an accuracy estimate is provided as well: $A=1-E_\textrm{tot}/1733$. Even though the proposed method does not yield the least overall error when compared to literature at large, it is, however, the lowest IRMA error when using an \textbf{autoencoder} for dimensionality reduction. This is of particular strategic interest for a deep-learning approach when larger and more balanced data sets are available for medical image retrieval. The non-neural approaches include a handcrafted chain of processing (LBP plus saliency detection plus SVM classification, \cite{camlica2015medical}) and a dictionary approach with several fully customized preprocessing steps  \cite{avni2009addressing}.  

\begin{table*}[htb]
\centering 
\begin{tabular}{lcc}  
\toprule
Method  &  Error & Accuracy(\%) \\ \midrule
    Camlica et al. \cite{camlica2015medical}      & 146.55      & 91.54\%          \\ 
    Avni et al.  \cite{avni2009addressing}             & 169.50       & 90.22\%         \\ \hline\hline
\textbf{Autoencoder -- Proposed}                      &  \textbf{313.17}  &  81.93\%\\ 
Autoencoder -- Sze-To et. al \cite{sze2016binary}          & 344.08      & 80.14\%          \\ 
Autoencoder -- Sharma et. al \cite{sharma2016stacked}      & 376.00    & 78.30\%               \\ 
Autoencoder -- Tizhoosh et. al \cite{tizhoosh2016barcodes} & 392.09    & 77.37\%            \\ 
\bottomrule
\end{tabular}
\caption{IRMA results: The lower section compares methods using autoencoders. The upper section reports the best results by non-neural approaches. } 
\label{tab:IRMAresults}
\end{table*}

To summarize, we explored the use of autoencoders for retrieval of medical images. We employed three features, namely Radon projections, HOG features, and raw pixels. In order to provide a complete solution for image retrieval, we used an MLP for classification. When conducting experiments on the IRMA dataset, we observed that dimensionality reduction capabilities of autoencoder remove the redundancies of the input vector. We obtained the lowest  IRMA error of $313$ when autoencoders are employed. In addition, a comparative study between Radon projections, the histogram of oriented gradients (HOG), and raw pixels has been discussed which suggests that Radon projections, as an input feature vector for the autoencoder, perform better. Particularly, Radon projections are observed to perform faster compared to both raw images and HOG, while obtaining the lowest IRMA error in comparison. This is of special interest since Radon transform is a well-established method for medical imaging. The aforementioned IRMA error is the lowest IRMA error achieved when using autoencoders for dimensionality reduction and feature extraction. This is of interest because IRMA data set with its distinct imbalance poses a challenge to autoencoders that require, like many other neural networks, large and balanced datasets \cite{hinton2006reducing}. 

\bibliographystyle{IEEEtran}
\bibliography{references.bib}

\end{document}